\documentclass[11pt]{article}

\usepackage[final]{acl}

\usepackage{times}
\usepackage{latexsym}

\usepackage[T1]{fontenc}

\usepackage[utf8]{inputenc}

\usepackage{microtype}

\usepackage{inconsolata}

\usepackage{graphicx}

\usepackage{booktabs}
\usepackage{makecell}   
\usepackage{siunitx}    
\usepackage{multicol}
\usepackage{multirow}
\usepackage{adjustbox}
\usepackage{colortbl}       

\usepackage{float}
\usepackage{stfloats}

\usepackage{listings}
\usepackage{xcolor}
\lstdefinelanguage{yaml}{
  keywords={true,false,null,y,n},
  basicstyle=\ttfamily\footnotesize,
  comment=[l]{\#},
  morecomment=[s]{/*}{*/},
  stringstyle=\color{blue},
  morestring=[b]',
  morestring=[b]"
}

\title{Adapting Vision-Language Models for E-commerce Understanding at Scale}

\author{Matteo Nulli$^{1,2}$, Vladimir Orshulevich$^1$, Tala Bazazo$^1$, Christian Herold$^1$, \\ \textbf{Michael Kozielski$^1$, Marcin Mazur$^1$, Szymon Tuzel$^1$, Cees G. M. Snoek$^2$,}  \\ \textbf{Seyyed Hadi Hashemi$^1$, Omar Javed$^1$, Yannick Versley$^1$ and Shahram Khadivi$^1$} \\ $^1$eBay Inc.,  $^2$University of Amsterdam \\
  \texttt{\{mnulli, tbazazo\}@ebay.com}}

\begin{document}
\maketitle
\begin{abstract}

E-commerce product understanding demands by nature, strong multimodal comprehension from text, images, and structured attributes. General-purpose Vision–Language Models (VLMs) enable generalizable multimodal latent modeling, yet there is no documented, well-known strategy for adapting them to the attribute-centric, multi-image, and noisy nature of e-commerce data,  without sacrificing general performance. In this work, we show through a large-scale experimental study, how targeted adaptation of general VLMs can substantially improve e-commerce performance while preserving broad multimodal capabilities. Furthermore, we propose a novel extensive evaluation suite covering deep product understanding, strict instruction following, and dynamic attribute extraction.





\end{abstract}

\section{Introduction} \label{sec: Introduction}
Deep e‑commerce product understanding is inherently multimodal. 
While today's search works primarily through matching the textual part of a listing, images of an item, its packaging, or general visuals play a large role in how customers evaluate and select the item they want.
Recent advancements in Large Language Models (LLMs)
\citep{dubey2024llama3herdmodels, yang2024qwen2technicalreport, mistral2024small3}, have shown strong results on e-commerce tasks, with some specific approaches for domain-specific customization
\cite{peng2024ecellmgeneralizinglargelanguage, herold2025domainadaptationfoundationllms}. However, translating these gains into the vision–language setting, like we do in this paper, remains a considerable challenge.

\begin{figure}[t]
\centering
\includegraphics[width=\columnwidth]{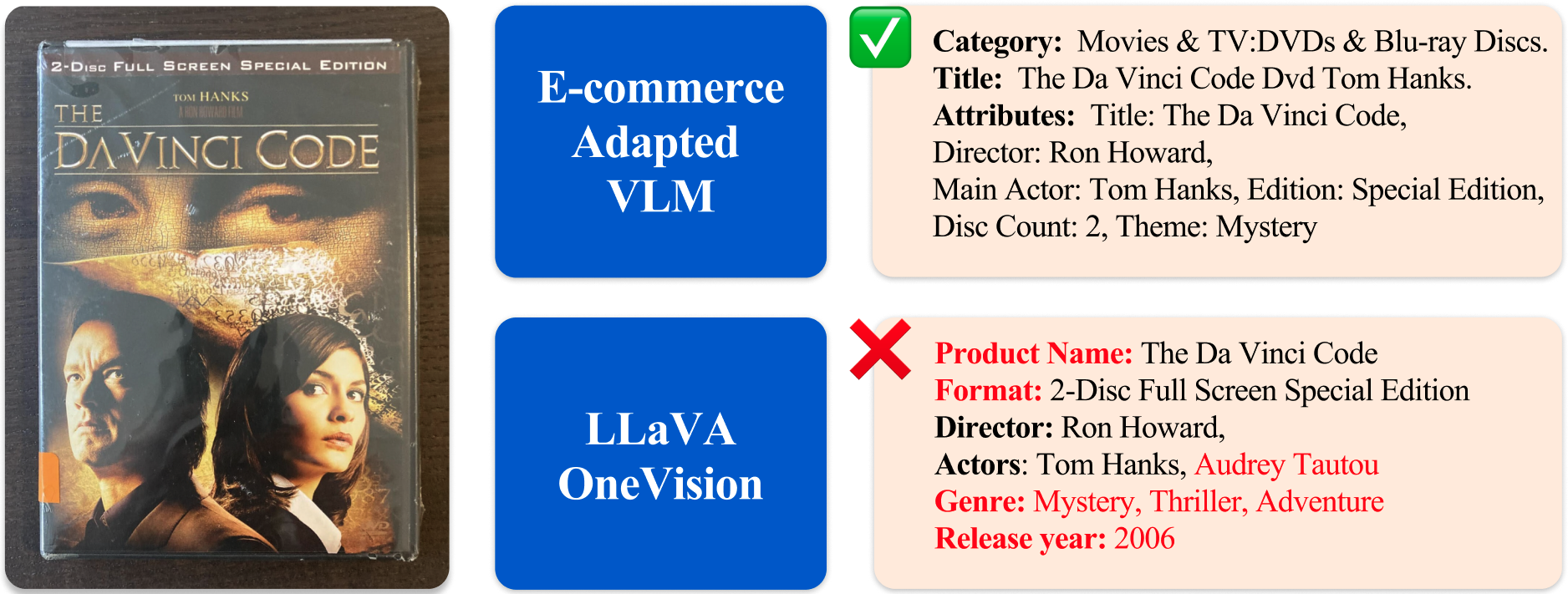} 
\caption{\textbf{Output of our E-commerce Adapted VLMs compared against same size LLaVA-OneVision.} We show our models ability to more faithfully extract attributes from e-commerce items. In \textcolor{red}{red}, we highlight wrong model predictions that are neither tied to the image nor valid item attributes. 
}
\label{fig:intro}
\end{figure}

General‑purpose Vision–Language Models (VLMs) such as LLaVA-OneVision \citep{li2024llavaonevisioneasyvisualtask}, Qwen3‑VL \citep{qwen3vl}, InternVL3 \citep{internvl2}, and Gemma3 \citep{gemmateam2025gemma3technicalreport}, have consistently achieved state-of-the-art results across a broad spectrum of downstream applications, encompassing image captioning \cite{yu2022coca, chen2023sharegpt4vimprovinglargemultimodal, wan2024loccavisualpretraininglocationaware}, visual question answering \cite{liu2024llavanext, li2024llavaonevisioneasyvisualtask}, deep image understanding \cite{tong2024cambrian1fullyopenvisioncentric,bai2025qwen25vltechnicalreport}, and complex reasoning tasks \cite{xu2024llavacotletvisionlanguage, nulli2025objectguided}, making the deployment of multimodal systems in e-commerce feasible. 
Nevertheless, we see a need for a reproducible, backbone-agnostic recipe for adapting VLMs to the demands of e‑commerce attribute‑centric reasoning, multi‑image aggregation, and robustness to noisy seller‑generated content, \emph{without} loosing general VLM-capabilities performance. 
Moreover, in spite of a large amount of evaluation sets for text‑only shopping tasks \cite{jin2024shoppingmmlumassivemultitask}, rigorous benchmarking of multimodal shopping assistants remains underdeveloped. 

In this paper, we focus on two questions, (i) if high‑performing e‑commerce VLMs truly require a customized LLM, or whether adapting on vision-focused tasks suffices. And (ii) on the best way to build a benchmark to assess multiple dimensions of understanding from extracting product attributes to category-specific deeper understanding and handling of multi-image tasks. 
To tackle (i) we perform \textbf{extensive ablations across multiple visual and text decoders} as backbones. Moreover, we propose a new set of \textbf{multimodal instruction data} to strengthen e-commerce abilities without hindering general performance, showing adaptation is possible.  
To answer (ii), we propose a set of benchmarks evaluating a broad range of \textbf{internal use-cases and real-life online retail} scenarios. In summary our contributions are as follows:
\begin{itemize}
    \item We show how to \textbf{adapt existing VLMs towards the e-commerce domain}, taking into account task-specific features, and demonstrate it enhances performance on online shopping tasks considerably, without any loss of capabilities on other domains.
    \item We design and implement a comprehensive set of vision, \textbf{e-commerce benchmarks} based on real production problem statements and data.
    \item We also evaluate state-of-the-art VLMs across general-domain and in-domain multimodal tasks, reporting our adaptation findings across data mixtures, models sizes and architectures. 
\end{itemize}

All in all we provide insights, evaluation suites and a proven strategy for an e-commerce adaptation of VLMs, retaining strong general capabilities. 

\section{Related Work} \label{sec: related work}

\paragraph{e-Commerce Vision Language Models}
Online shopping platforms such as eBay own an enormous quantity of data which can be leveraged when training LLMs and VLMs. Among the many applications, the ability of models to concretely grasp user-uploaded \emph{visual}-information, correctly comprehending multimodal product characteristics and being able to predict them accordingly are vital features in online marketplace applications. Research efforts such as \citet{bai2023qwenvlversatilevisionlanguagemodel, xue2024pumgptlargevisionlanguagemodel, li2024multimodalincontexttuningapproach}, finetune VLMs for product understanding and tackle product description generation exploiting in-context learning capabilities.
Similar e-commerce adaptation works like \citet{ ling2024captionsspeaklouderimages} instruction tune Llama-3.2 model with online shopping data. 
While these are interesting research directions, none have yet concurrently studied the effect of multiple pre-trained multimodal architectures on downstream online retail performance, all while being able to retain effectiveness on general purpose multimodal benchmarks. 

\paragraph{E-commerce-specific Evaluation}
Text‑centric suites \cite{jin2024shoppingmmlumassivemultitask}
have helped standardize measurement of general shop‑assistant abilities and even powered community competitions, but they operate primarily on textual signals. Similar widely used datasets evaluate query–product relevance, review‑grounded product Q\&A, purchase‑intention comprehension and domain factuality via knowledge graphs \cite{reddy2022shopping, gupta2019amazonqareviewbasedquestionanswering, ding2024intentionqabenchmarkevaluatingpurchase, chen2025chineseecomqascalableecommerceconcept, liu2025eckgbenchbenchmarkinglargelanguage}.  
While general‑purpose VLM evaluations \cite{fu2024mmecomprehensiveevaluationbenchmark} stress broader visual‑language understanding, like visual-question answering or object recognition, they are not tailored to the e-commerce fine‑grained attributes and tool use typical of retail. In recent research, \citet{ling2025ecommmmu} covers some question answering, product classification and relevance-related tasks as well as product relation identification and sentiment analysis and their dataset, while large-scale and comprehensive, is built by taking text-only datasets, adding images and removing the image-text pairs where the images are redundant, whereas we feel that our setting of taking image-focused tasks as a starting point is more naturalistic.

\section{Methodology} \label{sec: methodology}

\begin{figure*}[h]
\centering
\includegraphics[width=\textwidth]{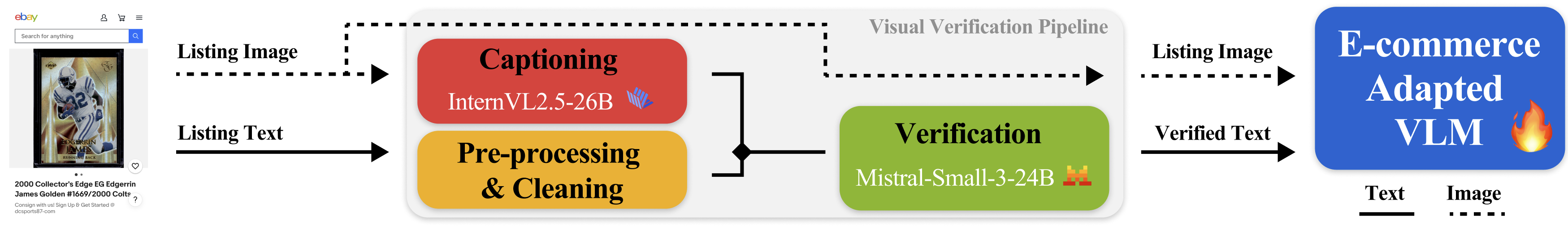}
\vspace{-2em}
\caption{\textbf{Visual Verification Pipeline.} The figure shows the pipeline we use to create the 4M e-commerce visual instruction tuning data. We begin by collecting raw listings data from the web (\textit{left}). We then clean and pre-process the textual entries. In parallel, we create detailed \textit{captions} for the corresponding image through InternVL-2.5-26B. Finally, we provide the \textit{captions} together with the \textit{cleaned listings} to Mistral-Small-3-24B to obtain the \textit{verified} instructions, used, along with original images, to train our models (shown with fire).}
\label{fig: data-curation}
\end{figure*}

\subsection{Our E-commerce Benchmarks} \label{sec: ecommerce evaluation}

To tackle the gap in multimodal e-commerce-specific benchmarks, we propose a set of four evaluation suites described below. Each is designed to tailor 
internal eBay production use-cases, ranging on a variety of tasks, categories and metrics.

\paragraph{Aspect Prediction}
Our Aspect Prediction evaluation set, divided into three different sub-parts. The first, comprised of 2600 general questions on all e-commerce categories, and the second two, evaluate the model's ability to predict aspects in Fashion, with and without additional contexts from item title and category, both with 1600 examples. All are evaluated through string matching.

\paragraph{Deep Fashion Understanding}
We design a specialized benchmark consisting of 3000 samples divided into three subsets: \emph{Apparel Men Shirts and Women Tops}, \emph{Handbags}, and \emph{Sneakers}. Each subset targets critical attributes relevant to the product type, structured into clear classification categories. Evaluation involves prompting the model to categorize items precisely according to the provided attribute classes.

\paragraph{Dynamic Attribute Extraction}
This evaluation set comprises 1,000 synthetically generated with GPT-4o \cite{gpt4o}, human‑verified examples. It benchmarks a model’s ability to enumerate and structure all visually grounded attributes from an image without a predefined schema. 

\paragraph{Multi-image Item Intelligence}
In this dataset the model is asked to compile a fixed set of attributes related to compliance questions (e.g.\ brand, warning labels, ingredients) from multiple product items into a structured  \texttt{JSON} output, enabling verification and recall matching processes. 1000 items were sampled to prioritize product categories with high regulatory requirements (toys, electronics, electrical appliances, cosmetics, etc.). We evaluate through LLM-as-a-judge
(see e.g.\ \citealp{gu2025surveyllmasajudge}).
More on each set in Appendix \ref{app: sec: ecommerce_benchmarks}.

\vspace{-2pt}
\subsection{Our Approach to E-commerce Adaptation} \label{sec: ecommerce training}
\vspace{-2pt}
We first go through our Data Curation pipeline, VLM Adaptation Training Stages, additional Multi-Image Item Intelligence specific fine-tuning and the architectures on which we apply this adaptation. 

\subsubsection{Internal Data Curation}

Raw e-commerce listings data is typically rather noisy, containing redundant and incomplete information or just simply wrong inputs. Yet high-quality data is crucial when training large multimodal models. Here, we show how to leverage the self-supervised signal inherent in user-generated listings data and describe our \textit{Visual Verification Pipeline} for large-scale data curation, illustrated in Figure \ref{fig: data-curation}.
We begin by collecting nearly 15 million raw listings from online marketplace websites and select only the primary (main) image for each listing. Each image is captioned through InternVL-2.5-26B \cite{chen2025expandingperformanceboundariesopensource}. Alongside, we extract the user-supplied item aspects from each listing. Given the generated caption and item aspects, we employ Mistral-Small-3-24B \cite{mistral2024small3} to verify which of these aspects can be inferred from the caption and thus from the image itself. This verification ensures visual-textual correspondence during training. 

The resulting listings, enriched with the verified aspects and paired with their original images, form the high-quality dataset used to train our multimodal models.

\subsubsection{General E-commerce Adaptation}
\label{subsubsec:general_domain_train}
Following LLaVA‑OneVision~\cite{li2024llavaonevisioneasyvisualtask}, we train our models in three stages: 
(i) Vision-Language Alignment, (ii) Mid‑Stage Training, and (iii) Visual Instruction Tuning.
For (i) we employ LLaVA-OneVision set of instructions with 
BLIP‑LAION 558k corpus  \cite{liu2023visualinstructiontuning} and for (ii) their \href{https://github.com/LLaVA-VL/LLaVA-NeXT/blob/main/scripts/train/mid_stage.yaml}{mid‑stage mixture} \cite{li2024llavaonevisioneasyvisualtask} removing several subsets that we found low‑signal or redundant. 

\paragraph{Visual Instruction Tuning} 
\label{sec: visual_instruction_tuning}
Finally, we conduct instruction tuning on (a) a version of the LLaVA‑OneVision \href{https://github.com/LLaVA-VL/LLaVA-NeXT/blob/main/scripts/train/single_image.yaml}{single-image mixture}, and (b) $\sim$4M internal e‑commerce oriented set of instructions pictured in Appendix Figure \ref{fig: ebay-si-data}. This portion is partitioned as follows, with percentages equaling part of e‑commerce total:
\textbf{VQA} (45\%), consists of free‑form, yes/no, image‑only questions, full item description all with and without title~\& category context. \textbf{Dynamic Attribute Extraction} (30\%), containing free‑form visual attribute extraction with and without title~\& category context. Variants include augmenting it with OCR, prompt constraining text, and any combinations of these settings. \textbf{Precise Instruction Following} (12.5\%), a set of keyword‑conditioned instructions that require inclusion/avoidance of specific terms and tasks emphasizing strict form/length control. \textbf{Listings} (12.5\%), comprised of full product listings predictions from an image. Details in Appendix~\ref{app: sec: ecommerce training}.

\begin{figure*}[!t]
\begin{center}
\includegraphics[width=\textwidth]{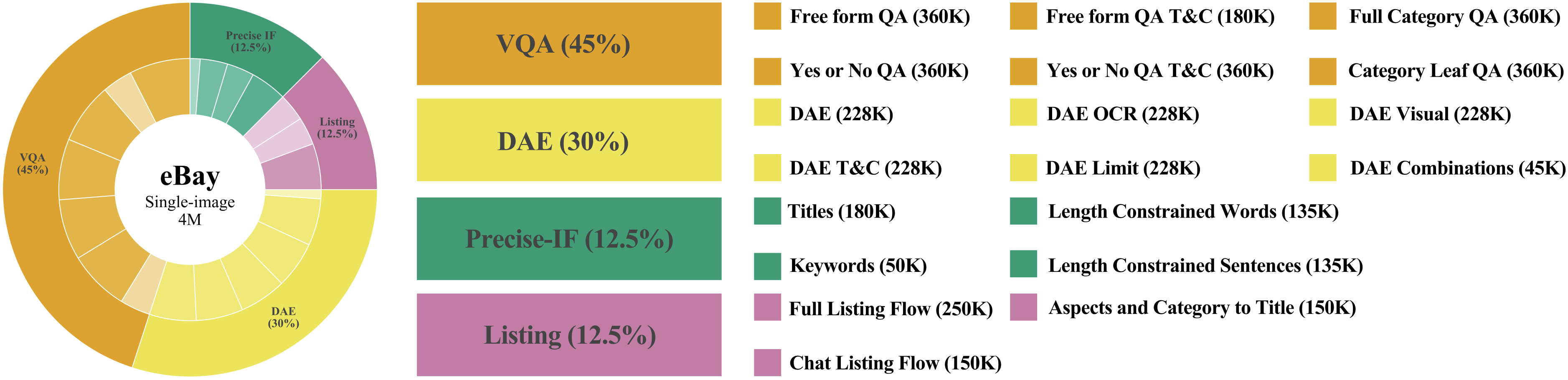}
\end{center}
\vspace{-0.3cm}
\caption{\textbf{eBay Single-Image Visual Instruction Tuning Set.} We break down the components of our internal single-image instruction tuning set. The pie chart on the left shows the percentages of tasks in our set. On the right we breakdown each tasks with its own sub tasks with the total number of instructions in parenthesis.}
\label{fig: ebay-si-data}
\end{figure*}

\subsubsection{Item Intelligence Fine-Tuning}

For our internal production Multi-Image Item Intelligence task, we curate a fine-tuning dataset of 100,000 items across relevant categories, each containing multiple images (median = 5, range = 2–8). Since no labeled data is available, we generate first annotations using GPT-4.1 via prompt-engineering. We then enhance the quality of both teacher annotations and inference-time inputs to focus on visually and semantically informative regions --- often textual or numeric details on product surfaces. We achieve this employing Qwen2.5-VL-32B \cite{bai2025qwen25vltechnicalreport} to produce precise bounding boxes, which are post-processed (expanded and merged) for better coverage. Cropped regions and original images are then re-annotated by GPT-4.1, yielding substantially higher-quality \emph{better labels}. More details in Appendix \ref{app: sec: item intelligence crops}.

\label{subsubsec:I2_train}
\vspace{-4pt}
\subsubsection{Model Architectures} \label{sec: architecture}

We compare several state-of-the-art (SOTA) model components for our e-commerce VLM.
For the vision encoder, we experiment with \textbf{SigLIP2-SO400M-Patch14-384} \cite{tschannen2025siglip2multilingualvisionlanguage} and \textbf{Qwen2.5 ViT} \cite{bai2025qwen25vltechnicalreport}. As text decoder, we compare \textbf{Llama3.1-8B} \cite{touvron2023llama}, \textbf{e-Llama3.1-8B} \cite{herold2025domainadaptationfoundationllms} an e-commerce adapted version of Llama3.1 8B, \textbf{Lilium 1B/4B/8B} \cite{herold2024liliumebayslargelanguage} trained from scratch for the e-commerce domain and \textbf{Qwen3 4B/8B} \cite{yang2025qwen3technicalreport}. Furthermore, we also adapt fully fledged SOTA VLMs for certain tasks, namely
\textbf{\href{https://huggingface.co/nvidia/Llama-3.1-Nemotron-Nano-VL-8B-V1}{Llama-3.1-Nemotron-Nano-VL-8B-V1}}, \textbf{Gemma3 4B/12B/27B} \cite{gemmateam2025gemma3technicalreport}, \textbf{Qwen2.5VL-7B} \cite{bai2025qwen25vltechnicalreport} and \textbf{Qwen3VL-8B} \cite{qwen3vl}. 
 
\newcounter{modelrow}

\begin{table*}[htbp]
\centering
\begin{adjustbox}{max width=\textwidth}
\begin{tabular}{
l
S[table-format=2.2]
S[table-format=2.2]
S[table-format=2.2]
S[table-format=2.2]
S[table-format=2.2]
S[table-format=2.2]
}
\toprule
\multirow{2}{*}{\textbf{Vision Encoder | LLM}} 
& \multicolumn{3}{c}{\textbf{Aspect Prediction}} 
& \multicolumn{2}{c}{\textbf{Deep Fashion Understanding}} 
& \multicolumn{1}{c}{\textbf{Dynamic Attribute Extraction}} \\
\cmidrule(lr){2-4} \cmidrule(lr){5-6} \cmidrule(lr){7-7}
& {General} 
& {Fashion} 
& {Fashion + T\&C} 
& {Apparel} 
& {Sneakers \& Handbags} 
& {DAE} \\
\midrule
 \rowcolor[gray]{0.9} \textbf{Internal E-commerce Adaptation} & & & & & & \\
$^\text{\tiny \stepcounter{modelrow}\themodelrow}$ {SigLIP2 | Llama-3.1-8B} & 37.7 & 46.0 & 51.9 & 67.0 & 75.1 & 59.7 \\
$^\text{\tiny \stepcounter{modelrow}\themodelrow}$ {SigLIP2 | e-Llama3.1-8B} & 44.4 & 52.8 & 60.4 & 78.9 & 79.5 & 66.1 \\
$^\text{\tiny \stepcounter{modelrow}\themodelrow}$ {Qwen2.5ViT | e-Llama3.1-8B} & 53.3 & 55.1 & 65.3 & 71.0 & 70.1 & \bfseries 70.7 \\
$^\text{\tiny \stepcounter{modelrow}\themodelrow}$ {SigLIP2 | Qwen-3-4B} & 54.6 &  \bfseries 60.7 & 67.5 & 78.6 & 80.1 & 66.5 \\
$^\text{\tiny \stepcounter{modelrow}\themodelrow}$ {SigLIP2 | Qwen-3-8B} & \bfseries 56.2 & 60.1 & \bfseries 68.5 & \bfseries 79.8 & \bfseries 81.6 & 68.1 \\
$^\text{\tiny \stepcounter{modelrow}\themodelrow}$ {SigLIP2 | Lilium-1B} & 41.0 & 48.4 & 54.4 & 72.2 & 71.0 & 66.3 \\
$^\text{\tiny \stepcounter{modelrow}\themodelrow}$ {SigLIP2 | Lilium-4B} & 42.3 & 49.1 & 56.7 & 74.7 & 73.5 & 68.3 \\
$^\text{\tiny \stepcounter{modelrow}\themodelrow}$ {SigLIP2 | Lilium-8B} & 42.4 & 49.2 & 55.8 & 75.2 & 77.0 & 68.0 \\
$^\text{\tiny \stepcounter{modelrow}\themodelrow}$ SigLIP | Gemma3-4B & 54.8 & 58.3 & 67.0 & 78.6 & 80.3 & 67.6  \\
 \rowcolor[gray]{0.9} \textbf{Open Source} & & & & & & \\
$^\text{\tiny \stepcounter{modelrow}\themodelrow}$ SigLIP | Qwen2-7B\,\,\,\,\,\,\,\,\,\,\,\,\,\,\,\,\,\,\textit{LLaVA-OV} & 28.7 & 30.3 & 47.4 & 62.8 & 39.5 & 67.0 \\
$^\text{\tiny \stepcounter{modelrow}\themodelrow}$ Qwen2.5ViT | Qwen2-7B\,\,\,\,\textit{Qwen2.5-VL} & 36.9 & 36.8 & 47.7 & 82.9 & 80.6 & 72.0\\
$^\text{\tiny \stepcounter{modelrow}\themodelrow}$ Qwen3ViT | Qwen3-8B\,\,\,\,\,\,\,\,\,\textit{Qwen3-VL} & \bfseries 40.5 & \bfseries 42.4 & \bfseries 58.2 & \bfseries 84.3 & \bfseries 84.6 & \bfseries 70.9 \\
$^\text{\tiny \stepcounter{modelrow}\themodelrow}$ SigLIP | Gemma3-4B\,\,\,\,\,\,\,\,\,\,\,\,\,\,\textit{Gemma3} & 24.3 & 29.0 & 40.4 & 64.2 & 77.5 & 72.7  \\
\bottomrule
\end{tabular}
\end{adjustbox}
\vspace{-4pt}
\caption{\textbf{Internal tasks comparison across model architectures and sizes.} We report performance of vision encoder and LLM combinations on three of our proposed evaluation sets (top row). "Internal E-commerce Adaptation" models indicate VLMs fully trained top to bottom starting from pre-trained backbones, "Open Source" indicates models not trained by us, the original \textit{model names} are next to their architectural structure. Higher is better ($\uparrow$). 
}
\label{tab: ellama-llama-ebay}
\end{table*}

\begin{table*}[h]
\centering
\begin{adjustbox}{max width=\textwidth}
\begin{tabular}{
l
S[table-format=2.2]
S[table-format=4.2]
S[table-format=3.2]
S[table-format=2.2]
S[table-format=2.2]
S[table-format=2.2]
S[table-format=2.2]
S[table-format=2.2]
S[table-format=2.2]
}
\toprule
\multirow{2}{*}{\textbf{Vision Encoder | LLM}} 
& \multicolumn{4}{c}{\textbf{Multimodal General Understanding}} 
& \multicolumn{1}{c}{\textbf{Vision}} 
& \multicolumn{2}{c}{\textbf{OCR, Chat/Doc QA}} 
& \multicolumn{1}{c}{\textbf{Reasoning}}
& \multicolumn{1}{c}{\textbf{e-Commerce}} \\
\cmidrule(lr){2-5} \cmidrule(lr){6-6} \cmidrule(lr){7-8} \cmidrule(lr){9-9} \cmidrule(lr){10-10} 
& {MMBench} 
& {MME} 
& {MME} 
& {MMStar} 
& {CVBench} 
& {TextVQA} 
& {AI2D} 
& {MMMU} 
& {eComMMMU} \\
& {(dev)} 
& {(Perc.)} 
& {(Cogn.)} 
& {} 
& {} 
& {(val)} 
& {(val)} 
& {(val)}
& {(test)} \\
\midrule
\rowcolor[gray]{0.9} \textbf{Internal E-commerce Adaptation} & & & & & & & & & \\
$^\text{\tiny \stepcounter{modelrow}\themodelrow}$ {SigLIP2 | Llama-3.1-8B} & 75.8 & 1556.1 & 314.6 & 49.5 & 62.3 & 75.2 & 76.3 & 43.9 & 46.9 \\
$^\text{\tiny \stepcounter{modelrow}\themodelrow}$ {SigLIP2 | e-Llama3.1-8B} & 76.9 & 1549.1 & 379.3 & 52.6 & 72.7 & 74.0 & 78.2 & 42.0 & 52.2\\
$^\text{\tiny \stepcounter{modelrow}\themodelrow}$ {Qwen2.5ViT | e-Llama3.1-8B} & 71.7 & 905.8 & 333.2 & 53.6 & 61.6 & 65.2 & 76.6 & 39.7 & 55.4 \\
$^\text{\tiny \stepcounter{modelrow}\themodelrow}$ SigLIP2 | Qwen-3-4B  & 81.0 & 1623.0 & \bfseries 485.7 & 60.1 & 73.7 & 75.8 & 80.6 & \bfseries 50.4 & 20.9 \\
$^\text{\tiny \stepcounter{modelrow}\themodelrow}$ SigLIP2 | Qwen-3-8B & \bfseries 82.5 & \bfseries 1648.4 & 453.6 & \bfseries 62.2 & \bfseries 77.2 & \bfseries 77.7 & \bfseries 82.6 & 49.1  & 50.0 \\
$^\text{\tiny \stepcounter{modelrow}\themodelrow}$ SigLIP2 | Lilium-1B & 64.7 & 1383.5 & 278.9 & 39.0 & 57.4 & 66.4 & 63.9 & 35.4 & 48.6\\
$^\text{\tiny \stepcounter{modelrow}\themodelrow}$ SigLIP2 | Lilium-4B & 75.5 & 1484.8 & 334.6 & 47.1 & 61.8 & 69.7 & 74.8 & 37.8 & 46.5\\
$^\text{\tiny \stepcounter{modelrow}\themodelrow}$ SigLIP2 | Lilium-8B & 77.4 & 1439.2 & 335.4 & 51.4 & 71.4 & 71.5 & 76.9 & 42.3 & \bfseries 58.3\\
$^\text{\tiny \stepcounter{modelrow}\themodelrow}$ SigLIP | Gemma3-4B & 78.3 & 1617.9 & 433.2 & 54.9 & 69.8 & 76.6 & 80.7 & 43.8 & 45.4 \\
\rowcolor[gray]{0.9} \textbf{Open Source} & & & & & & & & &\\
$^\text{\tiny \stepcounter{modelrow}\themodelrow}$ SigLIP | Qwen2-7B\,\,\,\,\,\,\,\,\,\,\,\,\,\,\,\,\,\,\textit{LLaVA-OV} & 76.4 & 1537.4 & 439.6 & 55.4 & 27.9 & 71.0 & 80.0 & 46.4 & \bfseries 50.8 \\
$^\text{\tiny \stepcounter{modelrow}\themodelrow}$ Qwen2.5ViT | Qwen2-7B\,\,\,\,\textit{Qwen2.5-VL} & 81.9 & 1677.7 & 654.6 & \bfseries 63.1 & \bfseries 32.8 & \bfseries 82.9 & 82.8 & 50.9 & 40.6 \\
$^\text{\tiny \stepcounter{modelrow}\themodelrow}$ Qwen3ViT | Qwen3-8B\,\,\,\,\,\,\,\,\,\textit{Qwen3-VL} & \bfseries 84.0 & \bfseries 1742.1 & \bfseries 660.7 & 62.2 & 26.6 & 80.9 & \bfseries 84.0 & \bfseries 52.4 & 47.6\\
$^\text{\tiny \stepcounter{modelrow}\themodelrow}$ SigLIP | Gemma3-4B\,\,\,\,\,\,\,\,\,\,\,\,\,\,\textit{Gemma3} & 67.9 & 1202.1 & 398.6 & 36.5 & 11.4 & 62.1 & 71.2 & 39.7 & 34.7 \\ 
\bottomrule
\end{tabular}
\end{adjustbox}
\vspace{-4pt}
\caption{\textbf{Public multimodal tasks comparison across model architectures and sizes.} We report performance of vision encoder and LLM combinations on public evaluation sets, we also show the split or metric in parenthesis (top row). "Internal E-commerce Adaptation" models indicate VLMs fully trained top to bottom starting from pre-trained backbones, "Open Source" indicates models not trained by us, the original \textit{model names} are next to their architectural structure. Higher is better ($\uparrow$).}
\label{tab: ellama-llama-gen}
\end{table*}

\section{Experiments} \label{sec: experiments} 
In our Experiments section, we compare our e-commerce adapted VLMs against existing ones (Section \ref{subsec:exp_other_vlms}), followed by an analysis of the importance of vision encoders (Section \ref{subsec:exp_vision_enc}) and text decoders (Section \ref{subsec:exp_text_dec}). In the second part, we focus on the  item intelligence use-case (Section \ref{subsec:exp_i2}).
\vspace{-6pt}
\subsection{Experimental Setup}
All models that we trained are optimized as described in Section \ref{sec: ecommerce training}. For training, we use the NeMo \cite{kuchaiev2019nemo} and LLaVA-OneVision frameworks \cite{li2024llavaonevisioneasyvisualtask}, using the same loss objective. Training was conducted on NVIDIA H100 GPUs (using up to 120 GPUs connected via NVLink and InfiniBand). In addition to our set of e-commerce benchmarks (see Section \ref{sec: ecommerce evaluation}), we also evaluate all models on a comprehensive set of public benchmarks.
We defer to the Appendix \ref{sec:mm_benchmarks} for a more detailed explanation of these sets.
\vspace{-8pt}
\subsection{Comparison against existing VLMs}
\label{subsec:exp_other_vlms}
We first compare our initial internally trained VLM \textbf{SigLIP2 | Llama-3.1-8B} against external VLMs as shown in Table \ref{tab: ellama-llama-gen} row 14 for general-domain benchmarks and in Table \ref{tab: ellama-llama-ebay} row 1 for e-commerce tasks.
We find that newer SOTA external VLMs like \textbf{Qwen3-VL-8B} outperform our internal model on the majority general-domain benchmarks.
However, on the e-commerce specific benchmarks, the picture is quite different. While some external models do perform very well on Deep Fashion Understanding, they do fall behind on most e-commerce specific benchmarks. This leads us to the conclusion that we need to invest in building our own customized VLM for relevant e-commerce tasks. In the following sections, we determine the best overall settings to accomplish this goal.
\vspace{-4pt}
\subsection{Importance of Vision Encoder}
\label{subsec:exp_vision_enc}
We begin this exploration by analyzing the importance of the vision encoder, comparing two architectures, \textbf{SigLIP2} and \textbf{Qwen2.5 ViT} while keeping the text encoder the same. On both e-commerce tasks (compare Table \ref{tab: ellama-llama-ebay} rows 2 \& 3), and general domain benchmarks (compare Table \ref{tab: ellama-llama-gen} rows 15 \& 16), the results are inconclusive, as there is no clear winner between the two encoders. This highlights the complicated relationship with the image modality and task definition, which we will also discuss below for the item intelligence task. For example, the native resolution feature of the \textit{Qwen2.5ViT} might be beneficial for tasks like aspect prediction, where small image details might be important, however we observe weaker results in more reasoning-oriented results in tasks like fashion understanding.
The gap between SigLIP2 \cite{tschannen2025siglip2multilingualvisionlanguage} and Qwen2.5ViT \cite{bai2025qwen25vltechnicalreport} is mostly apparent in high resolution scenarios, due to Qwen2.5ViT's ability to adapt to higher image sizes. The setting analyzed in both Tables shows benchmarks where images have low to mid resolutions. This largely decreases the performance enhancements of Qwen2.5ViT, leveling the playing field with respect to its counterpart. 

\subsection{Importance of Text-Decoder}
Comparing the impact of different LLMs when used as backbone with same vision encoder, we observe an influence of (a) domain knowledge of the LLM, (b) general knowledge and (c) model size, which we detail next.
\vspace{-4pt}
\label{subsec:exp_text_dec}
\paragraph{E-commerce Knowledge Helps}
\label{subsubsec:ecom_text_dec}
We compare VLMs based on \textbf{Llama-3.1 8B} against the \textbf{e-Llama3.1-8B} and \textbf{Lilium-8B} variants on the general-domain benchmarks (see Table \ref{tab: ellama-llama-gen} rows 14, 15, 21), with similar performance. This makes sense, as the underlying text-only LLMs do perform similar on general-domain text-based benchmarks as well. However, when looking at e-commerce specific performance (see Table \ref{tab: ellama-llama-ebay} rows 1, 2, 8) we find that the e-commerce knowledge of e-Llama and Lilium leads to a better adaptability. 
\vspace{-6pt}
\paragraph{General Capability Helps}
To see if and how the general-domain capabilities of the text decoder influence final performance, we compare \textbf{Qwen3} and \textbf{Gemma3} models against previous generation \textbf{(e)-Llama} and \textbf{Lilium}. 
The former are trained on significantly more data, therefore they exhibit higher performance on general domain text-only benchmarks. Generally, looking at Table \ref{tab: ellama-llama-gen}, and also comparing model sizes, we find that better capabilities of the text-decoder help improve performance on general domain VLM benchmarks. More interestingly, we find that they also lead to improvements on some e-commerce specific tasks (see Table \ref{tab: ellama-llama-ebay}), especially Aspect Prediction.
Together with the findings from Section \ref{subsubsec:ecom_text_dec}, this leads us to believe that further gains are possible using a domain-adapted version of the Qwen3/Gemma3 text-decoders, which we leave to future work.

\begin{table*}[htbp]
\centering
\begin{adjustbox}{max width=\textwidth}
\begin{tabular}{
l
S[table-format=2.1]
S[table-format=2.1]
S[table-format=2.1]
S[table-format=2.1]
S[table-format=2.1]
S[table-format=2.1]
}
\toprule
\multirow{2}{*}{\textbf{Model Name}} 
& \multicolumn{6}{c}{\textbf{Multi-Image Item Intelligence}} \\ 
\cmidrule(lr){2-7}
& {f1-score ($\uparrow$)} 
& {precision ($\uparrow$)} 
& {recall ($\uparrow$)} 
& {verifiable-correct ($\uparrow$)} 
& {verifiable-incorrect ($\downarrow$)}  
& {unverifiable ($\downarrow$)} \\
    \midrule
    \rowcolor[gray]{0.9} \textbf{0-shot} & & & & & & \\
    $^\text{\tiny \stepcounter{modelrow}\themodelrow}$  Gemma3 4B & 32.8 & 33.1 & \bfseries 36.7 & 53.6 & 21.3 & 25.1 \\
    $^\text{\tiny \stepcounter{modelrow}\themodelrow}$ Gemma3 27B \,\,\,\,\,\,\,\,\textit{primary image only} & 25.5 & 52.1 & 18.3 & 71.6 & 24.5 & 3.9 \\
    $^\text{\tiny \stepcounter{modelrow}\themodelrow}$ Gemma3 27B  & \bfseries 44.8 & \bfseries 61.8 & 36.6 & \bfseries 80.4 & \bfseries 15.9 & \bfseries 3.8 \\
    \rowcolor[gray]{0.9} \textbf{Finetuned} & & & & & & \\
    $^\text{\tiny \stepcounter{modelrow}\themodelrow}$ SigLIP2 | e-Llama3.1-8B & 42.5 & 57.0 & 35.3 & 72.0 & 24.0 & 4.0 \\
    $^\text{\tiny \stepcounter{modelrow}\themodelrow}$ Qwen2.5ViT | e-Llama3.1-8B & 28.7 & 60.4 & 20.4 & 72.2 & 26.0 & 1.9  \\
    $^\text{\tiny \stepcounter{modelrow}\themodelrow}$ Qwen2.5VL-7B  & 29.3 & 62.9 & 20.6 & 75.3 & 23.0 & 1.7  \\
    $^\text{\tiny \stepcounter{modelrow}\themodelrow}$ Llama-3.1-Nemotron-Nano-VL-8B-V1  & 50.9 & 63.3 & 44.0 & 79.2 & 18.9 & 1.9 \\
    $^\text{\tiny \stepcounter{modelrow}\themodelrow}$ Gemma3 4B  & 50.5 & 64.9 & 42.8 & 79.4 & 17.1 & 3.5 \\
    $^\text{\tiny \stepcounter{modelrow}\themodelrow}$ Gemma3 12B  & 51.8 & 67.7 & 43.5 & \bfseries 81.3 & 15.7 & \bfseries 3.1 \\
    $^\text{\tiny \stepcounter{modelrow}\themodelrow}$ Gemma3 27B  & \bfseries 52.6 & \bfseries 68.0 & \bfseries 44.6 & 81.2 & \bfseries 15.2 & 3.6 \\
    \rowcolor[gray]{0.9} \textbf{Finetuned with Better Labels} & & & & & & \\
    $^\text{\tiny \stepcounter{modelrow}\themodelrow}$ Gemma3 4B  & 53.8 & 68.1 & 49.6 & 82.7 & 15.9 & 2.0 \\
    $^\text{\tiny \stepcounter{modelrow}\themodelrow}$ Gemma3 12B  & 58.2 & 71.2 & 50.9 & 84.2 & 14.0 & 1.7 \\
    $^\text{\tiny \stepcounter{modelrow}\themodelrow}$ Gemma3 27B  & \bfseries 58.8 & \bfseries 71.0 & \bfseries 51.9 & \bfseries 85.2 & \bfseries 13.1 & \bfseries 1.6 \\
    $^\text{\tiny \stepcounter{modelrow}\themodelrow}$ Gemma3 4B  \,\,\,\,\,\,\,\,\textit{pan\&scan} & 56.9 & 68.3 & 50.5 & 83.1 & 15.1 & 1.8 \\
    $^\text{\tiny \stepcounter{modelrow}\themodelrow}$ Gemma3 4B  \,\,\,\,\,\,\,\,\textit{image crops} & 58.0 & 69.5 & 51.5 & 84.7 & 13.7 & \bfseries 1.6 \\
    \bottomrule
  \end{tabular}
  \end{adjustbox}
  \vspace{-4pt}
  \caption{\textbf{Multi-Image Item Intelligence Comparison.} We report performance of different models on multiple types of finetuning strategies (0-shot, Finetuned, Finetuned with Better Labels) over our multi-image item intelligence benchmark. The \textit{italic} next to the model names indicates different inference strategy.}
  \label{tab: item-intelligence}
\end{table*}

\vspace{-4pt}
\paragraph{Model Size: Important for Some Tasks}
Investigating the effect of the size of the text-decoder, we find a
consistent trend across both general-domain (Table \ref{tab: ellama-llama-gen}) and e-commerce-specific domain (Table \ref{tab: ellama-llama-ebay}).
In both cases, larger models lead to stronger performance.
However, there seems to be a task-depended threshold for which just increasing model size no longer seems to help.
For example, for the Fashion subset of the Aspect Prediction task, going from 1 billion to 4 billion parameters parameters leads to improvements, while going from 4 billion to 8 billion does not.
The latter is also consistent for both Lilium and the Qwen3 model families. A similar trend can be seen on MME.
We may attribute the lack of significant improvements across model sizes to the lack of task complexity. 

\subsection{Public E-commerce Benchmarking}
In the last column of Table \ref{tab: ellama-llama-gen} we report results on the Multi-Image E-comMMMU \cite{ling2025ecommmmu} benchmark. This set consists of 36.000 multi-image multitask understanding samples for e-commerce applications. Along with its relevance to this study, we decided to include this set as a control variable, un-biasing our considerations on our E-commerce Adaptation. 

\paragraph{E-commerce knowledge helps cross domains}
The difference between our Internal Adapted models and the Open Source ones is striking. It is clear how our adaptation delivers consistent results also on public e-commerce benchmarks, especially when comparing \textbf{Gemma3-4B} internal vs external (lines 22 and 26) with +11\% or lines 18 and 21 with 25 with +3\% and +11\% respectively. 

\paragraph{Adaptation generalizes to multi-images without training}
This increase is performance is even more impressive when considering our training set only consists of single-image instructions \ref{sec: visual_instruction_tuning}, compared to open models, trained on multi-image data. 

\paragraph{Decoder Size and Type are crucial}
Due to the multi-image nature of the benchmark, model size seems to be crucial, especially when comparing lines 17 with 18 and 19 and 20 with 21. Furthermore, employing previously trained e-commerce LLMs \cite{herold2024liliumebayslargelanguage, herold2025domainadaptationfoundationllms} results in a considerable performance boost, especially when comparing \textbf{SigLIP2 | Llama-3.1-8B} vs \textbf{e-Llama3.1-8B} and \textbf{Lilium-8B} with a 5\% and 12\% respective increase.
We defer to the Appendix \ref{app: sec: ecommmmu} with the full table of eComMMMU results per sub-task.

\subsection{Item Intelligence}
\label{subsec:exp_i2}
The Item Intelligence task extracts attributes targeted at regulatory compliance questions.
Our baseline is a non-customized Gemma3-27B. In our experiments, we show how we greatly improve quality and efficiency by fine-tuning on this task, while obtaining further improvements by modeling for task-specific characteristics.
\vspace{-4pt}
\paragraph{Single vs Multi-image}
\label{subsubsec:single_vs_multi}
We start by establishing the 0-shot performance of the \textbf{Gemma3-27B} VLM on the item intelligence task.
We compare two settings: (i) the model is given just the primary image of the corresponding listing (ii) the model is given the full set of images.
From Table \ref{tab: item-intelligence} row 28 \& 29, we can see that it is definitely beneficial for the model to have access to all existing images of a listing.
We also test the performance of the more efficient \textbf{Gemma3-4B} model (row 27), but find the model predictions to be of worse quality.

\vspace{-4pt}
\paragraph{Fine-Tuning Helps}
\label{subsubsec:finetuningI2}
Next, we compare fine-tuning a model and compare against the zero-shot approach from Section \ref{subsubsec:single_vs_multi}.
We fine-tune a subset of the models we discussed above for the general e-commerce adaptation.
As can be seen in Table \ref{tab: item-intelligence} row 36, fine-tuning significantly improves performance of the \textbf{Gemma3-27B} model.
Furthermore, performance of the much smaller \textbf{Gemma3-4B} VLM (row 34) is also strong after fine-tuning.
Other models like \textbf{Qwen2.5ViT | e-Llama3.1-8B} and \textbf{Qwen2.5VL-7B (ft)} fall behind.
Another big advantage of fine-tuning is the greatly improved inference efficiency.
Due to smaller model size and shorter prompt size, we achieve ca. 3.8x inference speedup when replacing Gemma3-27B with the finetuned Gemma3-4B model, while also improving on F1 Score, see Table \ref{tab:inference_speed} for results.

\begin{table}[]
\centering
\begin{adjustbox}{max width=\columnwidth}
\begin{tabular}{lcc}
\toprule
Model & {Sec/Example ($\downarrow$)} & {f1-score ($\uparrow$)} \\ \midrule
\rowcolor[gray]{0.9} \textbf{0-shot} & & \\
Gemma 27B & 25.5 & 44.8 \\
\rowcolor[gray]{0.9} \textbf{Finetuned} & & \\
Gemma 27B & 19.3 & \textbf{52.6} \\
Gemma 4B & \textbf{6.7} & 50.5 \\ \bottomrule
\end{tabular}
\end{adjustbox}
\caption{\textbf{Inference speed comparison.} We report the speeed comparison on the Multi-Image Item Intelligence task between the 0-shot Gemma 27B model and the 4B and 27B finetuned variants. We also report the f1-score from Table \ref{tab: item-intelligence}. Experiments were conducted on a single A100 GPU using a recent version of vLLM \cite{kwon2023efficient}.}
\label{tab:inference_speed}
\end{table}

\vspace{-4pt}
\paragraph{It Matters Where You Look}
\label{subsubsec:further_improvements}
In an effort to further improve results, we test the image bounding boxes approach outlined in Section \ref{subsubsec:I2_train}, which leads to better labels for training examples.
As can be seen from Table \ref{tab: item-intelligence} rows 37 - 39, this approach leads to significant improvements for all model sizes.
We also test including the image crops in inference (row 41) and compare against the \lq{}Pan \& Scan\rq{} feature from Gemma3 (row 40).
We find that both approaches improve performance, but our more targeted cropping leads to stronger results.

\vspace{-4pt}
\section{Conclusion} \label{sec: conclusion}
\vspace{-4pt}

We introduced a reproducible, backbone‑agnostic recipe for adapting open‑weight VLMs to the attribute‑centric, multi‑image, and noisy characteristics of e‑commerce. To evaluate this, we constructed a benchmark suite spanning {Aspect Prediction}, {Deep Fashion Understanding}, {Dynamic Attribute Extraction} and multi‑image {Item Intelligence}. Across extensive ablations we show how targeted adaptation can deliver substantial in‑domain gains while preserving broad capabilities and improving on out of distribution E-commerce data. 
Lastly, in a production‑style Item Intelligence case study, targeted cropping plus improved labels and fine‑tuning yielded strong quality gains and multiple times faster inference compared to general‑purpose VLMs.

\newpage
\section{Limitations}\label{sec: limitations}
Our study has the following limitations. 

\begin{itemize}
    \item \textbf{(i) Monolingual scope.} All model adaptation, supervision, and evaluation were conducted in English. Consequently, we do not characterize cross‑lingual transfer to product ontologies, attribute surface forms, or unit/size conventions that are language– and locale‑specific (i.e., multi‑script OCR for size charts, EU/JP sizing, or currency/decimal formats). 
 \item \textbf{(ii) Platform dependence.} The instruction corpus and benchmarks are sourced predominantly from a single marketplace, and many prompts/targets were curated or verified via automated pipelines. This creates potential distributional coupling to that platform’s taxonomy, seller conventions, imaging styles (studio vs. user‑generated), and metadata density. This hinders portability to other marketplaces with different attribute schema or listing norms remains uncertain. 
 \item \textbf{(iii) LLM‑mediated supervision and evaluation.} Portions of training signals (i.e., pseudo‑labels, instruction filtering) and some evaluations rely on LLMs. This introduces annotator bias, style bias, and measurement noise; moreover, evaluator–model family overlap can inflate or deflate measured gains due to inductive‑bias alignment in “LLM‑as‑judge” scenarios.
 
 \item \textbf{(iv) Coverage of phenomena.} While broad, our evaluation is not exhaustive: the Dynamic Attribute Extraction (DAE) set is $\sim\!1$k examples and category coverage emphasizes selected fashion and high‑volume verticals. As a result, performance on long‑tail categories, rare attributes, region‑specific variants, heavily composited images, or atypical listing styles is under‑constrained. Overall, the reported improvements should be interpreted as evidence of promise under these conditions rather than as guarantees of cross‑lingual or cross‑platform robustness.

\item \textbf{(v) Long Image Sequence Handling.} In scenarios with more than 10 images (rare),  we noticed our models may suffer from Out-of-Memory (OOM) issues as well as long inference times. This is particularly tricky for  Multi-Image Item Intelligence and eComMMMU benchmarks. While having 10 or more images is rare, this can lead to issues in potential production use-cases. While this could be solved by training larger context LLMs or through token efficient strategies \citep{zhang2025llavaminiefficientimagevideo}, it is something worth addressing in the future.
 \end{itemize}


\newpage
\bibliography{custom}

\newpage

\appendix

\section{Appendix}
\label{sec:appendix}

\subsection{Related Work (Continued)}
\label{sec:appendix_rel_work}

\paragraph{Multi Purpose MLLMs}
Since the advent of Visual Instruction Tuning \cite{liu2023visualinstructiontuning}, many have grasped the impact of combining CLIP Vision Encoders \cite{radford2021learningtransferablevisualmodels} with Large Language Models (LLMs) \cite{radford2019language, vicuna2023, touvron2023llama, dubey2024llama3herdmodels} to  enable cross modality understanding with LLMs. Most notably LLaVA \cite{liu2023visualinstructiontuning} and GPT4V \cite{openai2024gpt4technicalreport}, have paved the way for more diverse and varied MLLMs. Recent investigations have advanced along several complementary fronts. From a systematical decomposition of the training pipeline and characterization of model behavior across a variety of pre-trained backbones \cite{mckinzie2024mm1methodsanalysis, zhang2024mm15methodsanalysis, laurençon2024mattersbuildingvisionlanguagemodels} to the efficient processing of images spanning multiple resolutions \cite{liu2024llavanext, wang2024qwen2vlenhancingvisionlanguagemodels, internvl2} as well as the development of fully open multimodal foundation models \cite{deitke2024molmopixmoopenweights}.
Multimodal Large Language Models have consistently achieved state-of-the-art results across a broad spectrum of downstream applications, encompassing image captioning \cite{yu2022coca, chen2023sharegpt4vimprovinglargemultimodal, wan2024loccavisualpretraininglocationaware}, visual question answering \cite{liu2024llavanext}, image understanding \cite{liu2023visualinstructiontuning, tong2024cambrian1fullyopenvisioncentric}, and complex reasoning tasks \cite{xu2024llavacotletvisionlanguage, nulli2025objectguided}.


\paragraph{E-commerce Model Adaptation}
General-domain pretrained LLMs often struggle with domain-specific tasks, motivating domain-specific pretraining or targeted domain adaptation \cite{DBLP:conf/nips/LewkowyczADDMRS22, DBLP:journals/corr/abs-2311-16079, DBLP:journals/corr/abs-2308-12950}.

Pretraining a domain-specific LLM from scratch results in the highest degree of adaptation, including domain-specific knowledge, vocabulary, and more \cite{wu2023bloomberggpt, li2023starcoder, herold2024liliumebayslargelanguage}.
However, it is also extremely costly and slow, and requires a huge amount of domain-specific data.

As an alternative, continuous pretraining on in-domain text or fine-tuning an existing model can also substantially boost performance on domain-specific tasks \cite{DBLP:conf/iclr/AzerbayevSPSMJD24, DBLP:journals/corr/abs-2402-03300, DBLP:journals/corr/abs-2401-09646, herold2025domainadaptationfoundationllms}, at the cost of less overall customizability.

\paragraph{Vision Language Benchmarking}
The rapid evolution of VLMs has necessitated the development of rigorous benchmarking protocols to systematically assess model capabilities. Current evaluation pipelines extensively scrutinize performance across diverse cognitive and perceptual axes, including Image Reasoning \cite{chen2024rightwayevaluatinglarge}, Knowledge acquisition \cite{lu2022learn, lu2024mathvistaevaluatingmathematicalreasoning}, Perception \cite{ge2023planting}, and Vision-Centric analysis \cite{li2024naturalbenchevaluatingvisionlanguagemodels, tong2024cambrian1fullyopenvisioncentric}. While methodologies for assessing
Compositional Reasoning \cite{yuksekgonul2023visionlanguagemodelsbehavelike, nulli2024context}, Optical Character Recognition (OCR) \cite{singh2019vqamodelsread}, Science Reasoning \cite{lu2022learnexplainmultimodalreasoning} are becoming standardized \cite{yue2024mmmumassivemultidisciplinemultimodal, fu2024mmecomprehensiveevaluationbenchmark}, the process of evaluating e-Commerce related tasks—specifically Vision Question Answering for category attribution—remains undefined. We advocate for establishing a robust evaluation framework designed to rigorously measure Multimodal system performance within this specific domain.

\subsection{General Domain Multimodal Benchmarks} \label{sec:mm_benchmarks}
To evaluate our models on existing e-Commerce tasks we choose eComMMMU \cite{ling2025ecommmmu}, one of the few comparing evaluation suits for MLLMs in online shopping. It is comprised of over 35k multi-image samples spanning over 8 tasks. Furthermore, we employ 8 other general multimodal understanding benchmarks, ensuring close monitoring of general performance. These are MMBench \cite{liu2024mmbenchmultimodalmodelallaround} covering object detection, text recognition, action
recognition, among many others, MMMU \cite{yue2024mmmumassivemultidisciplinemultimodal} evaluating Mulimodal LLMs on perception, knowledge, and reasoning, CVBench \cite{tong2024cambrian1fullyopenvisioncentric} evaluating visual-centered capabilities of our models, and finally, MME \cite{fu2024mmecomprehensiveevaluationbenchmark}, a comprehensive benchmark dividing between perception and cognition tasks, with 15 subcategories. AI2D \cite{kembhavi2016diagramworthdozenimages} a Diagram/ChartQA with 3,009 examples, and MMStar \cite{chen2024rightwayevaluatinglarge} 1.5k samples across 6 categories (Perception, Math, Science \& Tech, Logical, Instance Reasoning). TextVQA \cite{singh2019vqamodelsread} designed to stress-test capabilities of VQA models in OCR, with 5k examples. 
Lastly, eComMMMU \cite{ling2025ecommmmu} consists of 36.000 multi-image multitask understanding samples for e-commerce applications and 8 sub-sets. This benchmark evaluates how MLLMs utilize visual information in real-world shopping scenarios.

\subsection{Methodology}
\label{sec:appendix_methodology}

\begin{figure}[t]
\centering
\includegraphics[width=0.98\columnwidth]{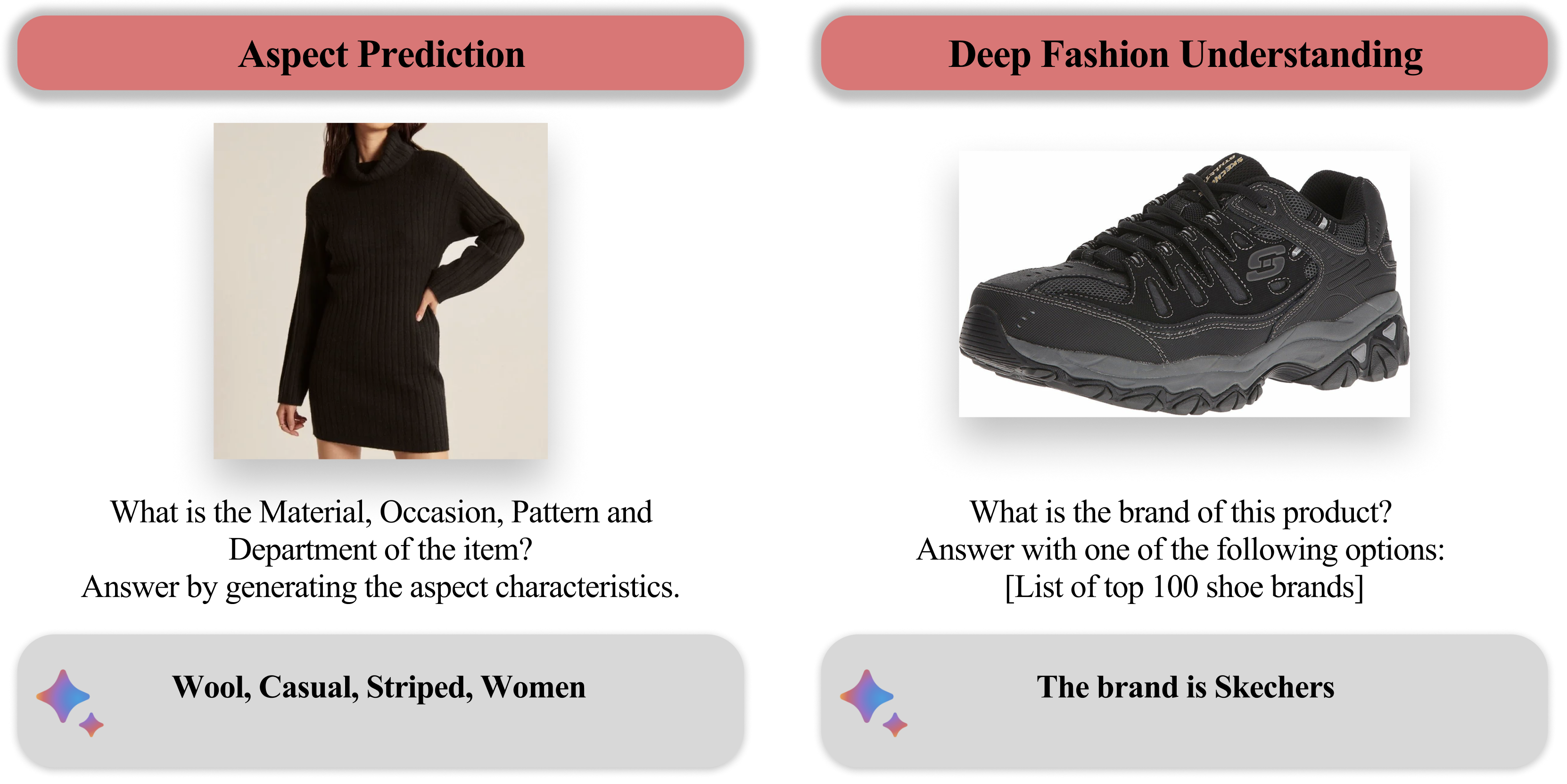}
\caption{\textbf{Benchmark examples from Aspect Prediction and Deep Fashion Understanding.} We choose a representative example from our Aspect Prediction and Deep Fashion Understanding benchmarks to showcase the tasks in detail.}
\label{fig: benchmarks-visualization}
\end{figure}

\begin{figure}[t]
\centering
\includegraphics[width=0.8\columnwidth]{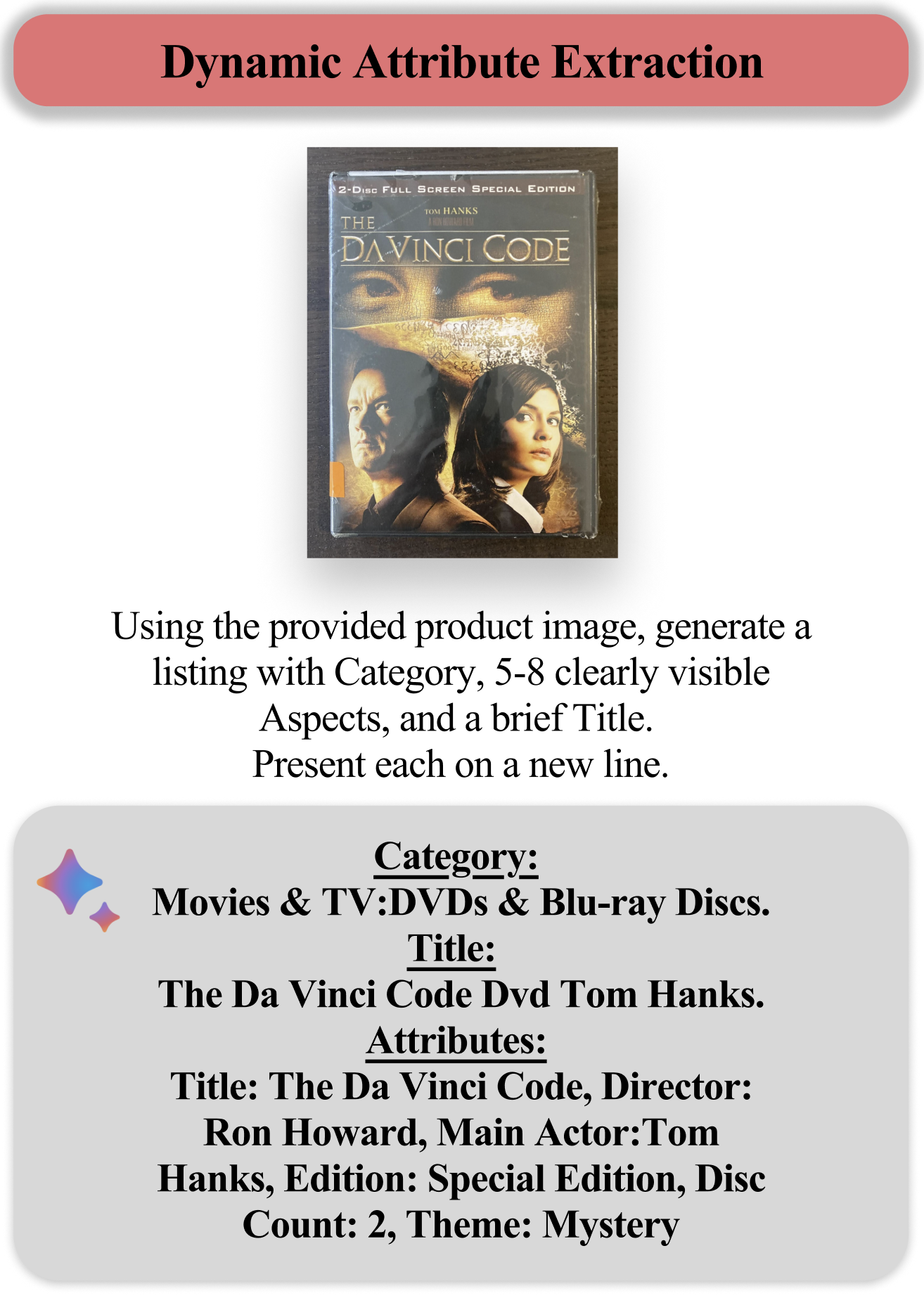}
\caption{\textbf{Benchmark example from Dynamic Attribute Extraction.} We choose a representative example from our Dynamic Attribute Extraction benchmark to showcase the task in detail.}
\label{fig: benchmarks-visualization2}
\end{figure}

\begin{figure}[t]
\centering
\includegraphics[width=0.8\columnwidth]{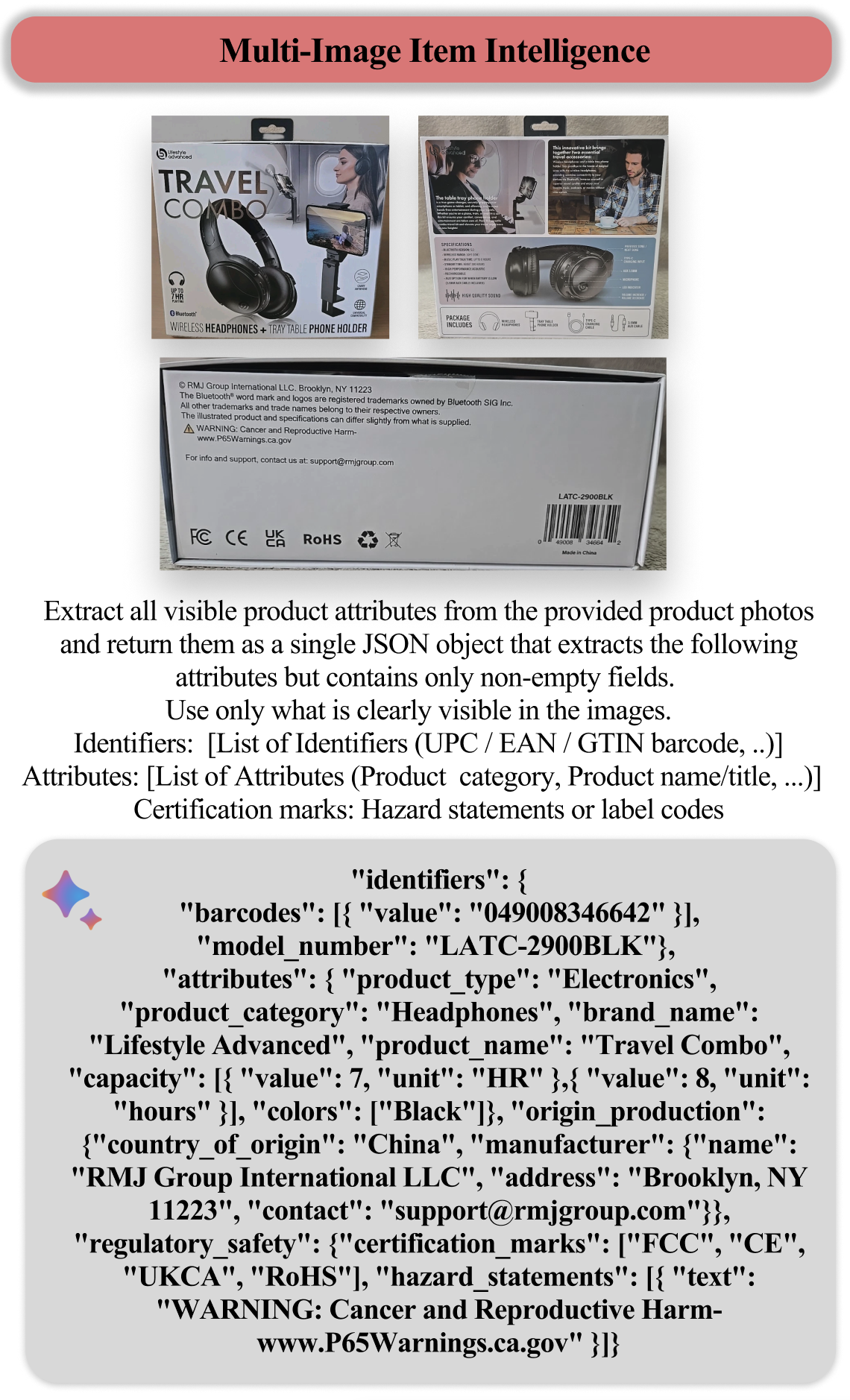}
\caption{\textbf{Benchmark example from Multi-Image Item Intelligence.} We choose a representative example from our Dynamic Attribute Extraction benchmark to showcase the task in detail.}
\label{fig: benchmarks-visualization3}
\end{figure}

\subsection{Our E-commerce Benchmarks} \label{app: sec: ecommerce_benchmarks}

\paragraph{Aspect Prediction}
We propose our Aspect Prediction evaluation suite. This set is divided into three different sub-parts, each tasked with a specific objective. 
The first set is comprised of 2600 general aspect prediction questions on
almost all e-commerce categories (collectibles, car parts, cards, fashion, etc...). In the last two, we evaluate the model's ability to predict aspects in Fashion, setting with and without additional textual contexts provided by item title and category, both with 1600 examples. All three are evaluated through string matching after post-processing. 
Although online shopping is often dominated by fashion items, we deem important to include evaluation sets which could more accurately capture the broad spectrum of online marketplaces.

\paragraph{Multi-image item intelligence}
Many attributes related to product safety and compliance such as certifications, ingredients, warning labels are not provided by the item's seller, and manual inspection
is inherently slow and costly. To address this, we propose a structured set designed to systematically extract and normalize visible information into consistent JSON outputs, enabling streamlined verification and recall matching processes. Our benchmark prioritizes product categories with prominent packaging and labeling signals, including toys, electronics, appliances, cosmetics, supplements, batteries, PPE, and food items. It handles diverse image sources such as product listing galleries, detailed zoomed-in views, and user-uploaded photographs. The resulting structured schema encompasses essential data elements such as \emph{Product Identifiers}, \emph{Product Attributes}, \emph{Product Origin}, and \emph{Regulatory Safety}, ensuring accurate and consistent outputs. We evaluate through LLM-as-a-judge. 

\paragraph{Deep Fashion Understanding}
Characterizing complex fashion features is a fundamental component of e-commerce assistants. To accurately evaluate deep fashion understanding, we designed a specialized sub-benchmark consisting of 3k samples divided into four distinct subsets: \emph{Apparel Men Shirts}, \emph{Apparel Women Tops}, \emph{Handbags}, and \emph{Sneakers}. Each subset targets critical attributes relevant to the product type, structured into clear classification categories. For instance, Apparel Men Shirts are evaluated based on Sleeve Length, Neckline, Pattern, and Color, with predefined classes such as 'Short Sleeve', 'Crew Neck', 'Striped', and 'Orange'. Apparel Women Tops share similar but more extensive attribute categories, including additional neckline and pattern options like 'Off the Shoulder' and 'Paisley'. Handbags and Sneakers subsets specifically focus on accurately identifying brand labels, such as 'Louis Vuitton' or 'Nike'. Evaluation involves prompting the model to categorize items precisely according to the provided attribute classes.

\paragraph{Dynamic Attribute Extraction}
Extracting visual item attributes from an image is a complicated yet essential task. This evaluation set benchmarks a model’s ability to enumerate and structure all visually grounded attributes from an image without a predefined schema. Each instance is prompted only once, requiring the model to decide which properties are salient, choose attribute names, and serialize values as key–value pairs (e.g., format, edition, material, artist, counts, genres, brand, model). The benchmark comprises 1,000 synthetically generated with GPT-4o \cite{gpt4o}, human‑verified examples and emphasizes attributes that are strictly supported by the pixels. Unlike fixed‑ontology extraction, Dynamic Attribute Extraction (DAE) stresses e-Commerce generalization by incentivizing exhaustive yet faithful outputs, avoiding hallucinated fields. A typical response for a text‑rich object, such as a DVD cover, would be a compact JSON record as show in Appendix \ref{fig: benchmarks-visualization}, \ref{fig: benchmarks-visualization2}, \ref{fig: benchmarks-visualization3}. By design, DAE probes the practical skill needed in cataloging, document understanding, and product intelligence workflows where schemas are fluid and attributes must be discovered on the fly.

\subsection{Item Intelligence Fine-tuning}
\label{app: sec: item intelligence crops}
Using both the original images and all derived crops for inference is computationally expensive, as the Gemma-3 image encoder assigns a fixed 256 visual tokens per image, causing inference cost to scale linearly with the number of images, even when many of them are small. On our training dataset, this resulted in a median of 12 and a maximum of 43 images per item. To address this, we construct crops covering the regions of interest optimized for the Gemma-3 encoder by identifying the smallest enclosing square that covers all bounding boxes, consistent with the model’s square image format. Finally, we apply a lightweight deduplication step using perceptual hashing (pHash) \cite{zauner2010phash}, reducing the number of images per item to a median of four and a maximum of nine. 

\subsection{Our Approach to E-commerce Adaptation} \label{app: sec: ecommerce training}

\begin{lstlisting}[language=yaml]
Our mid-stage datasets:
  - json_path: ./llava_ov/LLaVA-ReCap-558K.json
    sampling_strategy: all
  - json_path: ./llava_ov/LLaVA-ReCap-118K.json
    sampling_strategy: all
  - json_path: ./llava_ov/LLaVA-ReCap-CC3M.json
    sampling_strategy: all
  - json_path: ./llava_ov/synthdog_en_processed.json
    sampling_strategy: all
\end{lstlisting}

\begin{lstlisting}[language=yaml]
Our single-image LLaVA-OneVision sets for visual instruction tuning:
  - json_path: ./llava_ov/meta_ov/LLaVA-OneVision-Data_mavis_math_metagen.json
    sampling_strategy: "first:10%"
  - json_path: ./llava_ov/meta_ov/LLaVA-OneVision-Data_mavis_math_rule_geo.json
    sampling_strategy: "first:10%"
  - json_path: ./llava_ov/meta_ov/LLaVA-OneVision-Data_VisualWebInstruct(filtered).json
    sampling_strategy: "all"
  - json_path: ./llava_ov/meta_ov/LLaVA-OneVision-Data_chrome_writting.json
    sampling_strategy: "first:20%"
  - json_path: ./llava_ov/meta_ov/LLaVA-OneVision-Data_iiit5k.json
    sampling_strategy: "first:20%"
  - json_path: ./llava_ov/meta_ov/LLaVA-OneVision-Data_hme100k.json
    sampling_strategy: "first:10%"
  - json_path: ./llava_ov/meta_ov/LLaVA-OneVision-Data_orand_car_a.json
    sampling_strategy: "first:10%"
  - json_path: ./llava_ov/meta_ov/LLaVA-OneVision-Data_llavar_gpt4_20k.json
    sampling_strategy: "first:10%"
  - json_path: ./llava_ov/meta_ov/LLaVA-OneVision-Data_ai2d(gpt4v).json
    sampling_strategy: "all"
  - json_path: ./llava_ov/meta_ov/LLaVA-OneVision-Data_infographic_vqa.json
    sampling_strategy: "all"
  - json_path: ./llava_ov/meta_ov/LLaVA-OneVision-Data_infographic(gpt4v).json
    sampling_strategy: "all"
  - json_path: ./llava_ov/meta_ov/LLaVA-OneVision-Data_lrv_chart.json
    sampling_strategy: "all"
  - json_path: ./llava_ov/meta_ov/LLaVA-OneVision-Data_lrv_normal(filtered).json
    sampling_strategy: "first:10%"
  - json_path: ./llava_ov/meta_ov/LLaVA-OneVision-Data_scienceqa(nona_context).json
    sampling_strategy: "first:10%"
  - json_path: ./llava_ov/meta_ov/LLaVA-OneVision-Data_allava_instruct_vflan4v.json
    sampling_strategy: "first:30%"
  - json_path: ./llava_ov/meta_ov/LLaVA-OneVision-Data_allava_instruct_laion4v.json
    sampling_strategy: "first:30%"
  - json_path: ./llava_ov/meta_ov/LLaVA-OneVision-Data_textocr(gpt4v).json
    sampling_strategy: "first:10%"
  - json_path: ./llava_ov/meta_ov/LLaVA-OneVision-Data_ai2d(internvl).json
    sampling_strategy: "first:10%"
  - json_path: ./llava_ov/meta_ov/LLaVA-OneVision-Data_textcaps.json
    sampling_strategy: "first:10%"
  - json_path: ./llava_ov/meta_ov/LLaVA-OneVision-Data_ureader_cap.json
    sampling_strategy: "first:10%"
  - json_path: ./llava_ov/meta_ov/LLaVA-OneVision-Data_ureader_ie.json
    sampling_strategy: "first:10%"
  - json_path: ./llava_ov/meta_ov/LLaVA-OneVision-Data_vision_flan(filtered).json
    sampling_strategy: "all"
  - json_path: ./llava_ov/meta_ov/LLaVA-OneVision-Data_mathqa.json
    sampling_strategy: "all"
  - json_path: ./llava_ov/meta_ov/LLaVA-OneVision-Data_geo3k.json
    sampling_strategy: "all"
  - json_path: ./llava_ov/meta_ov/LLaVA-OneVision-Data_geo170k(qa).json
    sampling_strategy: "first:10%"
  - json_path: ./llava_ov/meta_ov/LLaVA-OneVision-Data_geo170k(align).json
    sampling_strategy: "first:10%"
  - json_path: ./llava_ov/meta_ov/LLaVA-OneVision-Data_sharegpt4o.json
    sampling_strategy: "all"
  - json_path: ./llava_ov/meta_ov/LLaVA-OneVision-Data_sharegpt4v(coco).json
    sampling_strategy: "all"
  - json_path: ./llava_ov/meta_ov/LLaVA-OneVision-Data_sharegpt4v(knowledge).json
    sampling_strategy: "all"
  - json_path: ./llava_ov/meta_ov/LLaVA-OneVision-Data_sharegpt4v(llava).json
    sampling_strategy: "all"
  - json_path: ./llava_ov/meta_ov/LLaVA-OneVision-Data_sharegpt4v(sam).json
    sampling_strategy: "all"
  - json_path: ./llava_ov/meta_ov/LLaVA-OneVision-Data_CLEVR-Math(MathV360K).json
    sampling_strategy: "first:10%"
  - json_path: ./llava_ov/meta_ov/LLaVA-OneVision-Data_FigureQA(MathV360K).json
    sampling_strategy: "first:10%"
  - json_path: ./llava_ov/meta_ov/LLaVA-OneVision-Data_Geometry3K(MathV360K).json
    sampling_strategy: "first:10%"
  - json_path: ./llava_ov/meta_ov/LLaVA-OneVision-Data_GeoQA+(MathV360K).json
    sampling_strategy: "first:10%"
  - json_path: ./llava_ov/meta_ov/LLaVA-OneVision-Data_GEOS(MathV360K).json
    sampling_strategy: "all"
  - json_path: ./llava_ov/meta_ov/LLaVA-OneVision-Data_IconQA(MathV360K).json
    sampling_strategy: "first:5%"
  - json_path: ./llava_ov/meta_ov/LLaVA-OneVision-Data_MapQA(MathV360K).json
    sampling_strategy: "first:10%"
  - json_path: ./llava_ov/meta_ov/LLaVA-OneVision-Data_PMC-VQA(MathV360K).json
    sampling_strategy: "first:1%"
  - json_path: ./llava_ov/meta_ov/LLaVA-OneVision-Data_Super-CLEVR(MathV360K).json
    sampling_strategy: "first:10%"
  - json_path: ./llava_ov/meta_ov/LLaVA-OneVision-Data_TabMWP(MathV360K).json
    sampling_strategy: "first:10%"
  - json_path: ./llava_ov/meta_ov/LLaVA-OneVision-Data_UniGeo(MathV360K).json
    sampling_strategy: "first:10%"
  - json_path: ./llava_ov/meta_ov/LLaVA-OneVision-Data_VizWiz(MathV360K).json
    sampling_strategy: "first:10%"
  - json_path: ./llava_ov/meta_ov/LLaVA-OneVision-Data_image_textualization(filtered).json
    sampling_strategy: "first:20%"
  - json_path: ./llava_ov/meta_ov/LLaVA-OneVision-Data_ai2d(cauldron,llava_format).json
    sampling_strategy: "all"
  - json_path: ./llava_ov/meta_ov/LLaVA-OneVision-Data_chart2text(cauldron).json
    sampling_strategy: "first:10%"
  - json_path: ./llava_ov/meta_ov/LLaVA-OneVision-Data_chartqa(cauldron,llava_format).json
    sampling_strategy: "first:10%"
  - json_path: ./llava_ov/meta_ov/LLaVA-OneVision-Data_diagram_image_to_text(cauldron).json
    sampling_strategy: "all"
  - json_path: ./llava_ov/meta_ov/LLaVA-OneVision-Data_hateful_memes(cauldron,llava_format).json
    sampling_strategy: "first:10%"
  - json_path: ./llava_ov/meta_ov/LLaVA-OneVision-Data_hitab(cauldron,llava_format).json
    sampling_strategy: "first:10%"
  - json_path: ./llava_ov/meta_ov/LLaVA-OneVision-Data_iam(cauldron).json
    sampling_strategy: "first:10%"
  - json_path: ./llava_ov/meta_ov/LLaVA-OneVision-Data_infographic_vqa_llava_format.json
    sampling_strategy: "first:10%"
  - json_path: ./llava_ov/meta_ov/LLaVA-OneVision-Data_intergps(cauldron,llava_format).json
    sampling_strategy: "first:10%"
  - json_path: ./llava_ov/meta_ov/LLaVA-OneVision-Data_mapqa(cauldron,llava_format).json
    sampling_strategy: "first:10%"
  - json_path: ./llava_ov/meta_ov/LLaVA-OneVision-Data_rendered_text(cauldron).json
    sampling_strategy: "first:10%"
  - json_path: ./llava_ov/meta_ov/LLaVA-OneVision-Data_robut_sqa(cauldron).json
    sampling_strategy: "first:10%"
  - json_path: ./llava_ov/meta_ov/LLaVA-OneVision-Data_robut_wikisql(cauldron).json
    sampling_strategy: "first:10%"
  - json_path: ./llava_ov/meta_ov/LLaVA-OneVision-Data_screen2words(cauldron).json
    sampling_strategy: "first:10%"
  - json_path: ./llava_ov/meta_ov/LLaVA-OneVision-Data_tabmwp(cauldron).json
    sampling_strategy: "first:5%"
  - json_path: ./llava_ov/meta_ov/LLaVA-OneVision-Data_tallyqa(cauldron,llava_format).json
    sampling_strategy: "first:5%"
  - json_path: ./llava_ov/meta_ov/LLaVA-OneVision-Data_st_vqa(cauldron,llava_format).json
    sampling_strategy: "first:10%"
  - json_path: ./llava_ov/meta_ov/LLaVA-OneVision-Data_visual7w(cauldron,llava_format).json
    sampling_strategy: "first:10%"
  - json_path: ./llava_ov/meta_ov/LLaVA-OneVision-Data_visualmrc(cauldron).json
    sampling_strategy: "first:10%"
  - json_path: ./llava_ov/meta_ov/LLaVA-OneVision-Data_vqarad(cauldron,llava_format).json
    sampling_strategy: "first:10%"
  - json_path: ./llava_ov/meta_ov/LLaVA-OneVision-Data_vsr(cauldron,llava_format).json
    sampling_strategy: "first:10%"
  - json_path: ./llava_ov/meta_ov/LLaVA-OneVision-Data_vistext(cauldron).json
    sampling_strategy: "first:10%"
  - json_path: ./llava_ov/meta_ov/LLaVA-OneVision-Data_websight(cauldron).json
    sampling_strategy: "first:10%"
    
\end{lstlisting}

\subsection{Experiments} \label{app: sec: Experiments}

\paragraph{eComMMMU} \label{app: sec: ecommmmu}
Given the similar goals of eComMMMU \cite{ling2025ecommmmu} and our work, we decided to include it within our general benchmarks. 
In Table \ref{tab: ecomMMMU} we show full results for eComMMMU on all 8 sub-tasks. 

We need to specify that we made some changes to (a.) the amount of images for each example and (b.) the final Average metric.
Regarding (a.) the eComMMMU paper uses either the main image or an (automatically) relevance-filtered subset which is not public. We first tried to include all images but hit Out-of-Memory issues. Some test-set examples contained north of 10 images. Due to our models context-sizes, we could not concurrently consider samples with these many images. Thus we capped the amount of images to 10 removing all excess, but keeping all textual examples. The second (b.) was a design choice on our side. We wanted to avoid to use the 'average model rank' for reproducibility and reporting purposes. We thus performed a weighted average across all tasks. This is what is shown in Table \ref{tab: ellama-llama-ebay} and as Avg. in Table \ref{tab: ecomMMMU}. 

\begin{table*}[htbp]
\centering
\sisetup{
  table-number-alignment = center,
  table-text-alignment = center,
  table-align-text-pre = false,
  table-align-text-post = false
}
\begin{adjustbox}{max width=\textwidth}
\begin{tabular}{
l
S[table-format=2.2]
S[table-format=2.2]
S[table-format=2.2]
S[table-format=2.2]
S[table-format=2.2]
S[table-format=2.2]
S[table-format=2.2]
S[table-format=2.2]
S[table-format=2.2]
}
\toprule
\multirow{2}{*}{\textbf{Vision Encoder | LLM}}
& \multicolumn{9}{c}{\textbf{eComMMMU (GTS)}} \\ 
\cmidrule(lr){2-10} 
&{AP} 
& {BQA} 
& {CP} 
& {SR} 
& {MPC}  
& {PSI}
& {SA}
& {PRP}
& {Avg.} \\
\cmidrule(lr){2-10} 
& \textit{Acc.} 
& \textit{Acc.} 
& \textit{Acc.} 
& \textit{R@1} 
& \textit{Acc.}  
& \textit{Acc.}
& \textit{Acc.}
& \textit{Acc.}
& \textit{}
\\
\midrule
\rowcolor[gray]{0.9} \textbf{Internal E-commerce Adaptation} & & & & & & & & & \\
$^\text{\tiny \stepcounter{modelrow}\themodelrow}$ {SigLIP2 | Llama-3.1-8B} & 66.6 & 33.6 & 49.8 & 5.9 & 64.0 & 27.8 & 50.1 & 31.0 & 46.9 \\
$^\text{\tiny \stepcounter{modelrow}\themodelrow}$ {SigLIP2 | e-Llama3.1-8B} & 33.6 & 17.8 & 50.5 & 5.7 & 64.0 & 68.5 & 70.9 & 50.2 & 52.5 \\
$^\text{\tiny \stepcounter{modelrow}\themodelrow}$ {Qwen2.5ViT | e-Llama3.1-8B} & 67.8 & 21.0 & 51.1 & 4.8 & 63.9 & 49.1 & 72.3 & 46.6 & 55.5 \\
$^\text{\tiny \stepcounter{modelrow}\themodelrow}$ SigLIP2 | Qwen-3-4B  & 1.0 & 1.0 & 32.4 & 0.0 & 63.0 & 6.4 & 4.8 & 38.5 & 20.9 \\
$^\text{\tiny \stepcounter{modelrow}\themodelrow}$ SigLIP2 | Qwen-3-8B & 65.2 & 34.4 & 50.8 & 7.9 & 65.1 & 33.2& 75.4 & 21.7 & 50.0 \\
$^\text{\tiny \stepcounter{modelrow}\themodelrow}$ SigLIP2 | Lilium-1B & 33.5 & 17.7 & 50.5 & 4.5 & 64.0 & 76.8 & 17.6 & 51.8 & 48.6\\
$^\text{\tiny \stepcounter{modelrow}\themodelrow}$ SigLIP2 | Lilium-4B & 34.0 & 18.0 & 50.4 & 4.6 & 44.6 & 76.6 & 57.9 & 28.5 & 46.5 \\
$^\text{\tiny \stepcounter{modelrow}\themodelrow}$ SigLIP2 | Lilium-8B & 59.0 & 31.8 & 50.4 & 4.6 & 64.0 & 73.2 & 70.9 & 39.3 & 58.3\\
$^\text{\tiny \stepcounter{modelrow}\themodelrow}$ SigLIP | Gemma3-4B & 65.4 & 33.2 & 51.9 & 6.7 & 64.0 & 24.7 & 58.9 & 14.5 & 45.5 \\
\rowcolor[gray]{0.9} \textbf{Open Source} & & & & & & & & &\\
$^\text{\tiny \stepcounter{modelrow}\themodelrow}$ SigLIP | Qwen2-7B\,\,\,\,\,\,\,\,\,\,\,\,\,\,\,\,\,\,\textit{LLaVA-OV} & 33.7 & 20.5 & 50.5 & 5.6 & 65.1 & 76.8 & 34.7 & 50.3 & 50.8 \\
$^\text{\tiny \stepcounter{modelrow}\themodelrow}$ Qwen2.5ViT | Qwen2-7B\,\,\,\,\textit{Qwen2.5-VL} & 31.2 & 46.2 & 32.5 & 10.0 & 65.7 & 26.9 & 58.0 & 37.0 & 40.6 \\
$^\text{\tiny \stepcounter{modelrow}\themodelrow}$ Qwen3ViT | Qwen3-8B\,\,\,\,\,\,\,\,\,\textit{Qwen3-VL} & 54.3 & 38.6 & 52.4 & 11.9 & 64.2 & 30.4 & 73.0 & 26.5 & 47.6 \\
$^\text{\tiny \stepcounter{modelrow}\themodelrow}$ SigLIP | Gemma3-4B\,\,\,\,\,\,\,\,\,\,\,\,\,\,\textit{Gemma3} & 45.2 & 32.5 & 50.3 & 11.0 & 39.7 & 29.9 & 49.0 & 14.6 & 34.7 \\ 
    \bottomrule
  \end{tabular}
  \end{adjustbox}
  \vspace{-4pt}
  \caption{\textbf{eComMMMU Full sub-tasks results.} We report performance of different models on \href{https://huggingface.co/datasets/NingLab/EcomMMMU}{eComMMMU test set} on the GTS subset with \emph{multiple} image per sample. We show performance on all sub-tasks ({AP = answerability prediction }, {BQA = binary question answering }, {CP = click through prediction}, 
{SR = sequential recommendation}, {MPC = multiclass product classification}, {PSI = production substitute identification}, {PRP = product relation prediction}, {SA = sentiment analysis}). For SR we report the Recall$@$1 score, whereas for all others accuracy. The Average (Avg) is calculated weighting based on the amount of samples per sub-task taking SR into account as well. The \textit{italic} next to the model names indicates different inference strategy.}
  \label{tab: ecomMMMU}
\end{table*}

\end{document}